# Tri-axial Self-Attention for Concurrent Activity Recognition

Yanyi Zhang
yz593@rutgers.edu

Xinyu Li
xxnl@amazon.com

Kaixiang Huang
hkx1@meitu.com

Yehan Wang
yw569@rutgers.edu

Shuhong Chen
sc1624@rutgers.edu

Ivan Marsic
marsic@soe.rutgers.edu

## Abstract

*We present a system for concurrent activity recognition. To extract features associated with different activities, we propose a feature-to-activity attention that maps the extracted global features to sub-features associated with individual activities. To model the temporal associations of individual activities, we propose a transformer-network encoder that models independent temporal associations for each activity. To make the concurrent activity prediction aware of the potential associations between activities, we propose self-attention with an association mask. Our system achieved state-of-the-art or comparable performance on three commonly used concurrent activity detection datasets. Our visualizations demonstrate that our system is able to locate the important spatial-temporal features for final decision making. We also showed that our system can be applied to general multilabel classification problems.*

1. Introduction

Research in human activity recognition has recently advanced from small-scale staged activities to large-scale real-world activities [16]. Attention and related strategies have helped the transition activity recognition from single-person with stationary backgrounds to multi-actor with moving backgrounds. Beyond single-image activity recognition, a new temporal feature extractor was proposed to achieve video-based activity recognition [16]. Concurrent activity recognition, however, has not been extensively researched. Existing research simply modifies single-activity recognition models with a sigmoid output layer or uses multiple single-activity recognizers for concurrent activity recognition; these have failed to achieve satisfactory performance [18].

Both concurrent and single-activity recognition require extraction of spatio-temporal features and associations, but the two tasks have many differences. Firstly, single-activity recognition focuses on selecting the most representative features associated with a certain activity; many attention-based methods were proposed to improve this feature extraction. Concurrent activity recognition, on the other hand, requires independent extraction of features for different activities. An unrelated feature to one activity might be critical for recognizing an associated activity. Secondly, single activity recognition focusses on modeling temporal feature associations from successive frames. This strategy does not work well for concurrent activity recognition, which requires separate temporal associations to be kept for different activities. Lastly, both types of activity recognition use probabilistic inference (softmax) for decision making. However, simply applying softmax with a threshold or using a parallel softmax layer for concurrent activity recognition would ignore the associations between activities.

To address these challenges, we introduce a concurrent activity recognition model with a feature-to-activity attention for feature extraction and a tri-axial self-attention encoder-decoder for multi-label prediction. The entire process is as follows. We first perform multi-level feature extraction, using VGG to first extract object-level features and then clustering them into global-level features containing information for all activities [10]. Instead of aggregating the extracted features in spatial and over time, we propose a feature-to-activity attention that maps the global features to sub-features associated with each activity. The proposed feature-to-activity attention maintains unaggregated temporal information to preserve temporal ordering; different activities may occur at different times and require different temporal attentions, which would be ignored by simple aggregation. We then propose a cascading self-attention structure that functions as an encoder-decoder. The encoder learns temporal attention for each activity type using a scaled dot-product. The decoder then generates concurrent activity predictions using another self-attention with our proposed activity association mask. Because three different dimensions (time, activity, and spatial features) are involved, we name the structure triaxial self-attention.

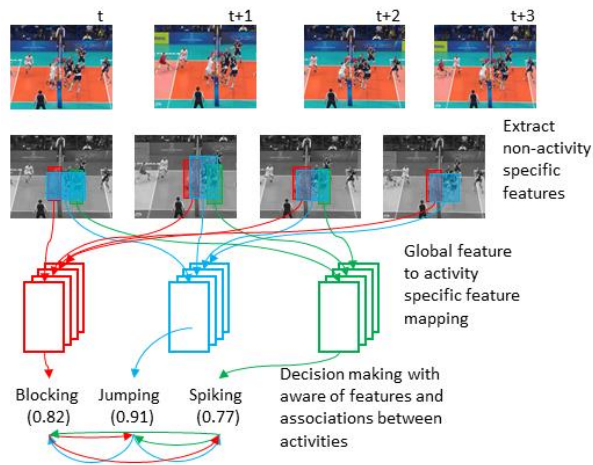

Figure 1: An overview of proposed concurrent activity recognition system that generates separated spatial-temporal features for independent activities.

We tested our system with published datasets: the hockey dataset [5] (12 activities), the volleyball dataset [15] (9 activities), and the charades dataset [25] (157 activities). We achieved state-of-the-art or comparable performance, and demonstrating that independently extracting spatio-temporal features for each activity helps with concurrent activity recognition. A visualization of our feature-to-activity mapping shows with triaxial self-attention, our system generates representative spatio-temporal features for each activity. Our contributions are:

- A three level feature extraction strategy with a novel feature-to-activity attention for concurrent activity recognition.
- A triaxial self-attention that learns independent temporal associations for different activities and makes concurrent activity predictions while being aware of possible activity combinations.
- An activity association mask that helps the self-attention-based decoder better capture the associations between activities.

The rest of this paper is organized as follows. Section 2 introduces related work. Section 3 describes our proposed method. Section 4 presents our experimental results on three published datasets. Section 5 discusses visualizations of key system components, as well as possible extensions and limitations. Section 6 concludes the paper.

2. Related Work

**Activity recognition** has been studied for decades. Traditional research was mainly based on hand-crafted features such as Histogram of Oriented Gradients (HOG) and Histogram of Optical Flow (HOF) [8]. Recently, deep learning has been commonly applied to activity recognition, initially using image recognition approaches with temporal feature fusion [16]. However, 2D CNNs originally used for image classification do not properly model spatio-temporal associations, which are critical for video-based activity recognition. To alleviate this issue, 3D convolution was proposed to learn both spatial and temporal features at once [7]. Although it was successful on some large activity-recognition datasets, 3D convolution is computationally expensive. CNN-LSTM strategies benefited from LSTM temporal association modeling. Spatial attention (or region-based methods) as well as temporal attention were proposed to help networks better focus on activity-associated features [24][19]. Since activity can usually be represented with a few key features, the idea of using clusters of features as activity descriptors was proposed in Action VLAD [10]. The iDT [31] and TDD [34] works showed that manually-crafted (as opposed to learned) spatio-temporal descriptors can achieve good activity recognition performance. Methods for single activity recognition assume that only one activity is contained in each video clip, which makes it possible to simply categorize the extracted features as related or unrelated to the activity. Concurrent activity recognition is significantly more challenging, as each video clip may contain an unknown number of activities.

**Concurrent activity recognition** has been researched only recently. Concurrent activities often occur in real-world scenarios such as sports games [5][15] or daily living scenarios [25]. Most existing approaches simply modify single-activity recognition strategies for concurrent activity recognition. A CNN-LSTM structure was tested for concurrent activity recognition with limited success [18]. More effective feature extraction frameworks were proposed in recent years, including a multi-stream network [14], a 3D convolution network [7][37], and a non-local neural network [32]. Objects and environmental information were used to help the activity recognition [26]. Instead of ordinary LSTMs, a new structure was proposed to better model temporal associations of activities [30][23]. But all of these strategies extracted global features shared for recognizing all the activities both spatially and temporally. Unlike previous approaches, we propose learning separate spatio-temporal associations for different activities instead of using global features for concurrent activity recognition.

3. Methodology

3.1. System Overview

Given a video ($T, W, H, 3$) with $T$ frames, our system does concurrent activity recognition in four steps: (1) **Feature extraction** (Figure 2: Feature Extraction) converts raw video frames into four dimensional features $F_{VGG} \in \mathbb{R}^{T \times W' \times H' \times C'}$ using a pre-trained VGG net. Features $F_{VGG}$ are then grouped into $K$ clusters of sub-features and further flattened into $F_{CF} \in \mathbb{R}^{T \times F}$, where $F$ denotes the feature dimension after flattening; (2) **Feature-to-activity attention** (Figure 2: Feature attention) maps the feature vector from each frame into a separate descriptor for each activity as $V \in \mathbb{R}^{T \times A \times F}$, where $T$ denotes the number of frames and $A$ denotes the number of activities. The flattened feature $F_{CF}$ only contains global features for all of extracted featues (Figure 2). Note that we keep the time information non-merged for now; (3) **Feature encoding** (Figure 2: Feature encoding) generates the temporal representation $F_A$ for each activity with scaled dot product [30]. $F_A \in \mathbb{R}^{A \times F}$ contains the spatio-temporal associations for all activities; (4) **Feature decoding** (Figure 2: Feature decoding) applies the multidimensional attention [23] to $F_A$ and generates a list of activity predictions based on both the encoded spatio-temporal features $F_A$ and associations between activities.

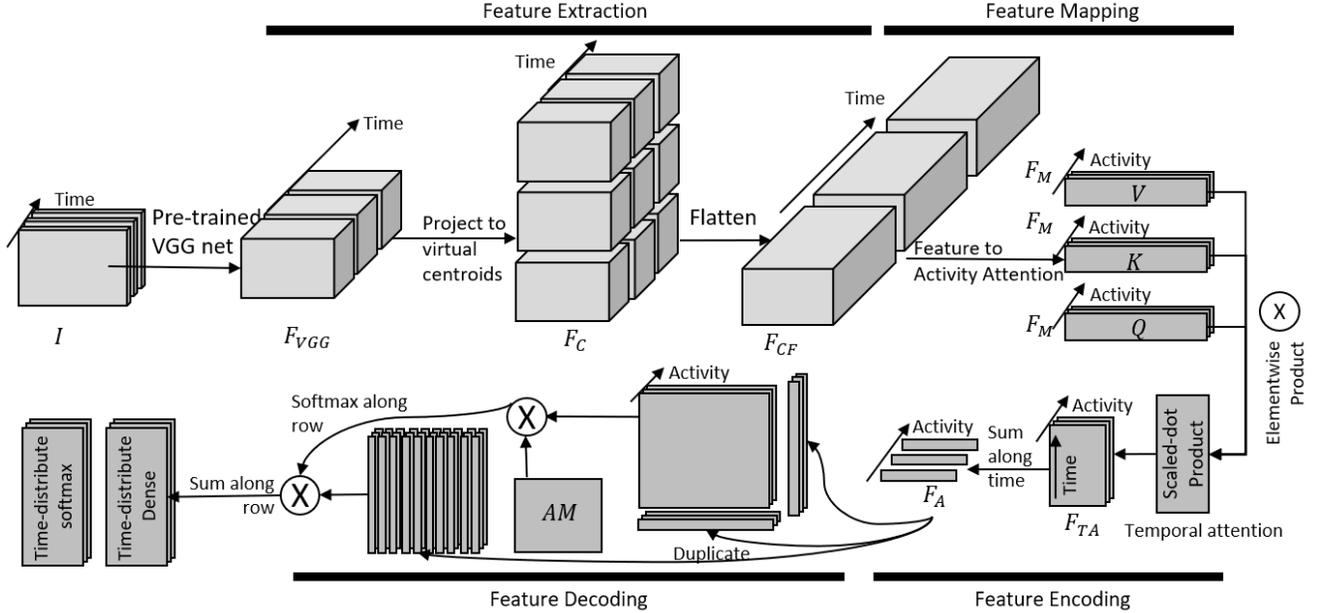

Figure 2: The proposed model for concurrent activity recognition using spatio-temporal attention.

### 3.2. Feature Extraction

We perform two-step feature extraction with VGG followed by feature clustering. The features of visual objects are first extracted by the pre-trained VGG net on the ImageNet dataset. Because single objects are not a good description for activity, we further combine several objects into a single descriptor, the activity cluster, for activity recognition. This is similar to the action VLAD [10].

We extracted the feature clusters by projecting the VGG features to K virtual cluster centroids (K=64 here, following previous work [10]). The centroids $c$ was first initialized using K-means and then fine-tuned during the network training:

$$V_t[i,j,k] = \frac{e^{-\alpha \|F_{VGG_{it}} - c_k\|^2}}{\sum_{k'} e^{-\alpha \|F_{VGG_{it}} - c_k\|^2}} (F_{VGG_{it}}[j] - c_k[j]) \quad (1)$$

where $F_{VGG} \in \mathbb{R}^{T \times W' \times H' \times C'}$ are the feature maps extracted from the pre-trained VGG16, $F_{VGG_{it}} \in \mathbb{R}^{C'}$ is the feature descriptor from $F_{VGG}$ at spatial location $i \in \{1,2,...,W \times H\}$ and at frame $t \in \{1,2,...,T\}$. $F_{VGG_{it}}[j]$ and $c_k[j]$ denote the $j$-th components of the feature descriptor $F_{VGG_{it}}$ and the $k$-th cluster centroid $c_k$. The difference ($F_{VGG_{it}}[j] - c_k[j]$) denotes the residual for the $j$-th component. The term $\frac{e^{-\alpha \|F_{VGG_{it}} - c_k\|^2}}{\sum_{k'} e^{-\alpha \|F_{VGG_{it}} - c_k\|^2}}$ denotes the soft-assignment and we set it to be tunable as proposed in [1].

We didn't directly aggregate the feature descriptors V as mentioned in [10] for two reasons: 1. Aggregating V across time ignores temporal ordering of the features. We confirmed this intuition by shuffling the frames in a video and feeding them into the trained VLAD network. The network generated the same softmax scores as for the original video. This experimental result demonstrates that Action VLAD cannot model the continuities of activities in temporal. 2. The activity cluster V contains important object-level features from the pretrained model. But with concurrent activities, each activity cluster may contain features associated with multiple activities (Figure 3, top right). Simply aggregating the descriptors from the activity cluster in spatial fails to extract representative features for each activity.

### 3.3. Activity Attention

Both activity cluster-based feature extraction and many other recently proposed methods use non-activity-specific ("global") features for concurrent activity recognition [10][7]. These features may contain information about multiple activities. For example, when multiple activities happen simultaneously in a hockey game (figure 3, top left), the activity cluster captures the features associated with all four activities (figure 3, top right). The global features are not representative for recognize each of the concurrent

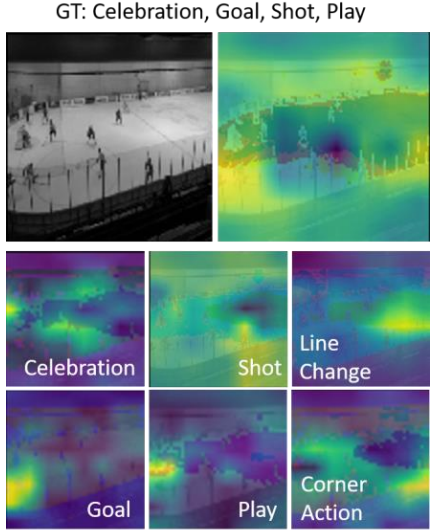

Figure 3: Attention for different activities (Celebration, Goal, Shot and Play are performed), top left the original video frame. Top right, the activation map of one activity cluster. Bottom, the generated attention for each activity by proposed feature to activity mapping.

activities.

In addition to the clusters of features, we propose a feature-to-activity mapping strategy that generates attention masks for each activity which, when applied to the global feature map, produce sub-features associated with each activity. This allows the model to use different features to recognize different activities. Consider $F_{CF} \in \mathbb{R}^{T \times W \times H \times F}$, the feature after flattening the VLAD descriptors from all clusters for the whole video clip (*T* frames per clip), where the dimension of *F* equals the number of clusters × feature vector length. We mapped $F_{CF}$ into *A* activity specific sub-spaces, where *A* is the number of activities. The feature to activity attention can be expressed as :

$$F_{Ma} = \sum_{w=1}^{W} \sum_{h=1}^{H} \mathcal{W}_a \otimes F_{CF} \qquad (2)$$

$$\mathcal{W}_a \in \mathbb{R}^{W \times H}, a \in \{1,2,\ldots,A\} \qquad (3)$$

where $W_a$ denotes the two-dimensional tunable parameters as an attention mask for activity *a*, which has the same shape with $F_{CF}$ it is multiplied with. $F_{Ma} \in \mathbb{R}^{T \times F}$ is the result after applying attention for activity a ($W_a$) onto $F_{CF}$ and aggregating the components in spatial. $F_M$ denotes the total set of $F_{Ma}$, with *a* from *1* to *A*. To affirm that the proposed mapping works, we visualized the activation maps (activation of $F_{CF}$) of independent activities after mapping (figure 3 bottom). The results show that proposed feature-to-activity mapping is able to map the global feature into features for each activity.

It is worth mentioning that we cannot simply set the number of centroids in VLAD to the number of activities for feature-to-activity mapping. This is because unsupervised learning methods can only learn features for low-level component actions and objects, which then in different combinations represent different high-level activities. Therefore, an additional network is necessary to learn the high-level activities as combinations of these features.

3.4. Triaxial self-attention encoder decoder

To learn different temporal associations for different activities is equally important as finding separate spatial features for different activities. This is because unlike single-activity recognition, multiple activities might be distributed through the entire video clip; the unspecific temporal attention will not be able to highlight the activity-specific features in time. Previous research modeled temporal features with a 3D ConvNet, LSTMs with attention, or 3D descriptors [18][7]. But all of these methods only extracted non-activity specific temporal associations. Additionally, many previous works simply used the top N results or a sigmoid layer for concurrent activity recognition, which ignores the possible associations between activities. Building the temporal associations for separate activities and making independent activity predictions with awareness of inter-activity associations is challenging. We propose a triaxial self-attention encoder that first encodes the activity-specific features over time into activity vectors and then makes a concurrent activity decoding from these encoded vectors.

*3.4.1 Temporal Encoding*

The key challenge for modeling temporal associations for different activities separately is to avoid unnecessary redundant parameters (e.g. having an independent LSTM encoder for each activity). The recently proposed transformer self-attention structure [30] can model independent sequential associations with a single network. The key for transformer self-attention is to extract separate sub-features with multi-head structure and then merge the information after applying scaled dot-product to different sub-features. Because our activity specific feature $F_M$ can be considered as equivalent to the features in multi-head attention after projected into several heads, we directly feed the activity specific feature into the scaled dot-product. In our implementation, we duplicated the $F_M$ for query, key and value required by the scaled dot-production, the scaled dot-product [30] is able to find the temporal attention vector $F_{Ta}$ for each activity as:

$$F_{Ta} = softmax\left(\frac{\mathcal{W}_a^q Q_a \mathcal{W}_a^k K_a^T}{\sqrt{F}} + Mask\right) \mathcal{W}_a^v V_a \qquad (4)$$

$$F_A = \sum_{t=1}^{T} F_{TA_t} \qquad (5)$$

$F_{Ta}$ denotes the result of the "Scaled Dot-product" for each activity *a*, which is the a-th component in the spatio-temporal feature for

| | waiting | setting | digging | falling | spiking | blocking | jumping | moving | standing |
|---|---|---|---|---|---|---|---|---|---|
| waiting | 1.00 | 0.28 | 0.43 | 0.16 | 0.41 | 0.54 | 0.09 | 0.61 | 1.00 |
| setting | 0.41 | 1.00 | 0.10 | 0.12 | 0.03 | 0.20 | 0.06 | 0.86 | 1.00 |
| digging | 0.44 | 0.07 | 1.00 | 0.16 | 0.24 | 0.40 | 0.06 | 0.51 | 1.00 |
| falling | 0.29 | 0.15 | 0.28 | 1.00 | 0.08 | 0.16 | 0.04 | 0.47 | 1.00 |
| spiking | 0.64 | 0.03 | 0.37 | 0.07 | 1.00 | 0.91 | 0.06 | 0.56 | 1.00 |
| blocking | 0.60 | 0.15 | 0.43 | 0.10 | 0.64 | 1.00 | 0.10 | 0.55 | 1.00 |
| jumping | 0.55 | 0.26 | 0.39 | 0.15 | 0.22 | 0.56 | 1.00 | 0.53 | 1.00 |
| moving | 0.44 | 0.42 | 0.36 | 0.19 | 0.26 | 0.36 | 0.06 | 1.00 | 1.00 |
| standing | 0.40 | 0.27 | 0.39 | 0.22 | 0.25 | 0.35 | 0.06 | 0.55 | 1.00 |

Positive activity association mask

| | waiting | setting | digging | falling | spiking | blocking | jumping | moving | standing |
|---|---|---|---|---|---|---|---|---|---|
| waiting | 0.00 | 0.72 | 0.57 | 0.84 | 0.59 | 0.46 | 0.91 | 0.39 | 0.00 |
| setting | 0.59 | 0.00 | 0.90 | 0.88 | 0.97 | 0.80 | 0.94 | 0.14 | 0.00 |
| digging | 0.56 | 0.93 | 0.00 | 0.84 | 0.76 | 0.60 | 0.94 | 0.49 | 0.00 |
| falling | 0.71 | 0.85 | 0.72 | 0.00 | 0.92 | 0.84 | 0.96 | 0.53 | 0.00 |
| spiking | 0.36 | 0.97 | 0.63 | 0.93 | 0.00 | 0.09 | 0.94 | 0.44 | 0.00 |
| blocking | 0.40 | 0.85 | 0.57 | 0.90 | 0.36 | 0.00 | 0.90 | 0.45 | 0.00 |
| jumping | 0.45 | 0.74 | 0.61 | 0.85 | 0.78 | 0.44 | 0.00 | 0.47 | 0.00 |
| moving | 0.56 | 0.58 | 0.64 | 0.81 | 0.74 | 0.64 | 0.94 | 0.00 | 0.00 |
| standing | 0.60 | 0.73 | 0.61 | 0.78 | 0.75 | 0.65 | 0.94 | 0.45 | 0.00 |

Negative activity association mask

Figure 4: The association mask for volleyball dataset.

all the activities $F_{TA} \in \mathbb{R}^{T*A*F}$. $F_A \in \mathbb{R}^{A*F}$ denotes the results after aggregating $F_{TA}$ along time. Instead of using position embedding, we applied forward and backward masks and as proposed in [23] for modeling the temporal order.

### 3.4.2 Activity decoding

The encoder part has coded the original input as a spatio-temporal feature representation for each activity ($F_a$). Decoding each activity needs to consider the feature associations with the other activities as well as considering the association between different activities based on statistic results. Previous decoder based on fully-connected layers with sigmoid [10][7] ignored the feature associations between different activities. Using an LSTM decoder for modeling the activity association [11] as still has two drawbacks: 1. LSTMs are time-consuming because of all the fully-connected operations for all the time instances, which requires $O(T \cdot F^2)$ per-layer. 2. An LSTM based assumes the activities were performed in order, but the activity outputs are not actually ordered. We choose to use multi-dimensional self-attention [23] instead of LSTM because self-attention was shown to have lower time complexity, which is $O(T^2 \cdot F)$ per-layer. Additionally, the "Scaled Dot-Product" method finds associations between every activity pair without considering activity order.

To model the association between activities, we propose statistic positive and negative activity association masks based on activity correlations in the training ground truth (Figure 4). The positive mask shows the co-occurrence of each activity pair:

$$Mask_{i,j}^{pos} = P(j = 1|i = 1) \quad (6)$$

where the mask is an $A \times A$ matrix, the $Mask_{i,j}^{pos}$ denotes the conditional probability that activity $j$ was performed given that activity $i$ was performed.

The negative mask shows the exclusion of each pair of activities. It is calculated as:

$$Mask_{i,j}^{neg} = 1 - Mask_{i,j}^{pos} \quad (7)$$

where $Mask_{i,j}^{pos}$ denotes the corresponding element in the positive mask. Note the negative masks that two activities are excluded with each other.

### 3.5. Implementation

We implemented our model in Keras with the tensorflow backend. We used the ReLU activation function for all convolutional operations and binary-cross entropy loss. We used the Adam optimizer with initial learning rate $10^{-4}$ and decreased it by factor of 10 after every 100 epochs. Batch normalization was used after each convolutional layer, and dropout (rate=0.5) was used after each dense layer to avoid overfitting. We trained our model with 4 Titan X GPUs and we trained each model for 1000 epochs.

Because all three concurrent activity recognition datasets have limited training samples, we performed spatio-temporal augmentation to avoid overfitting. We implemented the augmentation by first randomly down-sampling the frames (by a factor of 3) and then randomly selecting 64 consecutive frames. For frames selected from the same video, we performed the same random crop.

We used the Keras built-in VGG parameters for feature extraction, and set the last block as tunable. For the tunable VLAD descriptors, we made changes based on the action VLAD source code [36].

## 4. Experimental Results

### 4.1. Datasets

We tested our system with three commonly used concurrent activity recognition datasets:

***Hockey Dataset*** [5]: This was collected from real university-level hockey matches using two fixed cameras positioned at both ends of the rink on the spectator's side. It includes 36 videos with $480 \times 270$ resolution, sampled at 30 fps. We extracted the frames at 5 fps and randomly cropped the frame to $256 \times 256$ for training. Similar to previous works [28], we used 30 videos for training and 6 videos for testing. We set 15 as the size of sliding window for video sampling with 5 frames overlapping, as in [28].

***Volleyball Dataset*** [15]: This contains 55 videos with 4830 annotated frame dictionaries, 3493 for training and 1337 for testing. Each frame dictionary contains 41 frames at $1280 \times 720$ resolution. The label contains 9 types of concurrent activities. We down-sampled each frame dictionary to 18 frames and resized the frames to $256 \times 256$ due to hardware limitations. As multiple people are present in the volleyball game, we did not perform random cropping augmentations to avoid possible information loss.

***Charades Dataset*** [25]: This has 9848 videos of daily indoor activities showing 267 different users performing activities. The dataset includes 157 concurrent activities.

|  | Hockey Dataset | | Volleyball Dataset | |
| --- | --- | --- | --- | --- |
|  | Acc. | F1 | Acc. | F1 |
| VGG Net [27] | 91.1 | 25.2 | 50.2 | 37.5 |
| CNN-LSTM [18] | 92.10 | 37.40 | 53.10 | 38.80 |
| SVR [5] | / | 16.00 | / | / |
| EO-SVM [6] | 90.00 | / | / | / |
| CNN Overtime [28] | / | 42.0/61.0 | / | / |
| Hierarchical LSTM [29] | / | / | 72.70 | / |
| SRNN [3] | / | / | 76.65 | / |
| FCN+RNN [2] | / | / | 77.90 | / |
| Action VLAD [10] | 94.90 | 42.50 | 75.30 | 52.40 |
| Our Model | **96.10** | **50.50** | **79.40** | **64.50** |

Table 1: Experimental results on volleyball and hockey and comparison with previous work.

We used 7985 videos for training and the remaining 1863 for testing. We directly used the official 24 fps RGB frames. We performed the spatial-temporal augmentation as mentioned in section 3.5.

4.2. Results and Comparison

We compared our proposed method on the hockey and volleyball dataset with several baselines including: 1. AlexNet for framewise activity recognition. 2. CNN-LSTM with sigmoid layer [18]. We also compared our work with previous state-of-the-art works including: 1. Action VLAD (reimplemented based on official source codes [36]). 2. CNNs over time [28], and various SVMs. Based on our experimental results (Table 1, Figure 5) we found the proposed system outperformed Action VLAD and all the other state-of-the-art researches on both datasets. This is because of our proposed feature to action attention and triaxial self-attention would be able to extract individual spatio-temporal features for different activities and making activity predictions with aware of potential association between activities. We visualized the activity association map for results generated by our proposed model and some previous approaches (Figure 5). The results show that our model is able to better capture the reasonable association between activities. For example (Figure 5, red rectangle region), the jumping usually happened with blocking but jumping is hard to detect. Most of previous works is able to detect blocking but cannot associate it with jumping. It is worth to mention that [28] achieved higher F1 score than our proposed method, but they used weights adjustment strategy, which setting thresholding with arbitrary parameters. This strategy is not generalizable, because the model will be worse again if the testing sets are changed. Our methods outperformed their strategy without weights adjustment.

We also compared our method on the charades dataset (Table 2), the result shows our method outperformed most of the researches except the i3D and non-local neural network [32]. These two methods have much better performance on charades dataset, but they pre-trained their model on kinetics dataset (50 times larger) and the time complexity of their methods are 5 times larger than our proposed method. Due to the hardware limitation, we only trained our model on the charades dataset. Many researches [35] has demonstrated that with huge computational resources the simple model can have very good performance on large datasets. However, as discussed by previous work, have simple model running on limited resources with reasonable performance is also valuable. Our model achieved comparable performance with only 20% time complexity compared with i3D in charades. This is due to the proposed feature to activity attention and tri-axial self-attention which avoid using the large 3D convolutional layers.

| System | MAP | complexity (MACC) |
| --- | --- | --- |
| AlexNet [17] | 11.2 | 1.2 B |
| C3D[25] | 10.7 | 80 B* |
| Two-Stream + IDT [25] | 18.6 | / |
| CoViAR [35] | 21.9 | / |
| Asynchronous Temporal Fields [26] | 22.4 | / |
| i3D [7] | **34.4** | 165 B |
| Action VLAD [10] | 17.6 | 34 B |
| Non-local Neural Network [32] | **37.5** | 165 B+ |
| Our Model | 22.6 | 34 B |

Table 2: Experiment results on charades. MACC [9] (multiply and accumulate operations) is the measurement of time complexity. The complexity of Non-local neural network was written as 165 B+ because it was plugged into the original i3D, which having larger complexity than i3D. The complexity values with * means the author didn't propose the structure in detail, we estimated the result based on the similar method that the other proposed detailed structure.

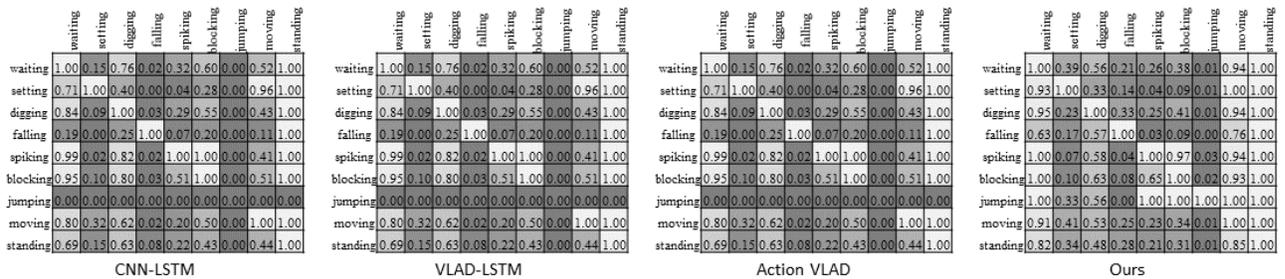

Figure 5: The generated activity association maps in volleyball dataset.

### 4.3. Structure Analysis

We further break down our proposed system and study the impact of each proposed component. Due to limited space, we only use hockey dataset examples. The experimental results demonstrated that (Table 3): 1. We tried to remove the proposed feature-to-activity attention by using the VLAD feature descriptor as [10] proposed but remaining unmerged-time. The system suffers from significant performance drops without our proposed feature-to-activity attention. This shows that mapping the global feature to each activity helps with concurrent activity recognition. 2. The system without activity clustering also has significantly lower performance, and we noticed that directly mapping features from VGG slows model convergence (5 times slower than using activity cluster). This is partially because directly mapping the object-level features to an activity is inefficient compared to first grouping useful features into a cluster. 3. We tried to replace the triaxial self-attention with a traditional LSTM encoder-decoder. The results show the proposed self-attention achieved better accuracy and efficiency: 8% higher F1 score with 50% less weights. The proposed model also converges 2 times faster than the LSTM model. 4. Finally, we replaced the proposed activity association mask with the normal forward/backward mask in the self-attention decoder. The experimental results show our proposed mask is able to help the system better model the association between different activities.

## 5. Discussion

### 5.1. Visualizing the triaxial encoder-decoder

We attempt to understand how the proposed triaxial self-attention models the temporal associations for different activities and makes the final prediction. We visualize the temporal attention for different activities learned by our system. We selected 15 frames of hockey video, with three activities (Shot, Save, and Play) happening simultaneously in last 5 frames. For comparison, we visualized the temporal attention for three concurrent activities and three activities that didn't happen in our selected clip (Checking, Fight and Goal). These are obtained from the output of $softmax\left(\frac{W_a^q Q_a W_a^k K_a^T}{\sqrt{F}} + Mask\right)$ in equation (4) in the network. The visualizations (Figure 6) show that the proposed self-attention strategy is able to generate different attention distributions for activities in progress and not in progress.

### 5.2. From Concurrent Activity Recognition to Multilabel Classification

Although we only demonstrated our proposed system on concurrent activity recognition, the core idea of feature-to-activity attention and tri-axial self-attention can be extended to general multi-label classification.

We tested our structure on the commonly used CelebA dataset [20]. We chose this dataset not only because it is a multi-label dataset, but also because it is a single-image multi-label classification dataset, which requires a slight modification of our system. Feature-to-activity attention can be refitted for multi-label classification problems by changing the dimension of $w_a$ from 3D to 2D (for static images); The triaxial self-attention can be scaled given different temporal dimensions and activity numbers. For a static image, the temporal dimension is 1, so there will be no temporal associations be found in the encoder part.

|  | Acc. | Prec. | Rec. | F1 |
|---|---|---|---|---|
| Without using feature cluster | 93.9 | 41.8 | 39.0 | 39.0 |
| Without feature to activity mapping | 95.2 | 48.9 | 43.6 | 44.5 |
| Replace the tri-axial self-attention with LSTM | 94.9 | 48.7 | 42.5 | 42.5 |
| Replace the self-attention-based decoder with LSTM | 94.6 | 49.3 | 45.0 | 43.5 |
| Replace AM with forward/backward mask | 95.4 | 50.3 | 44.4 | 45.7 |
| Complete model | **96.10** | **55.9** | **50.0** | **50.50** |

Table 3: Model structure analysis, evaluating the performance by removing each part of our proposed model.

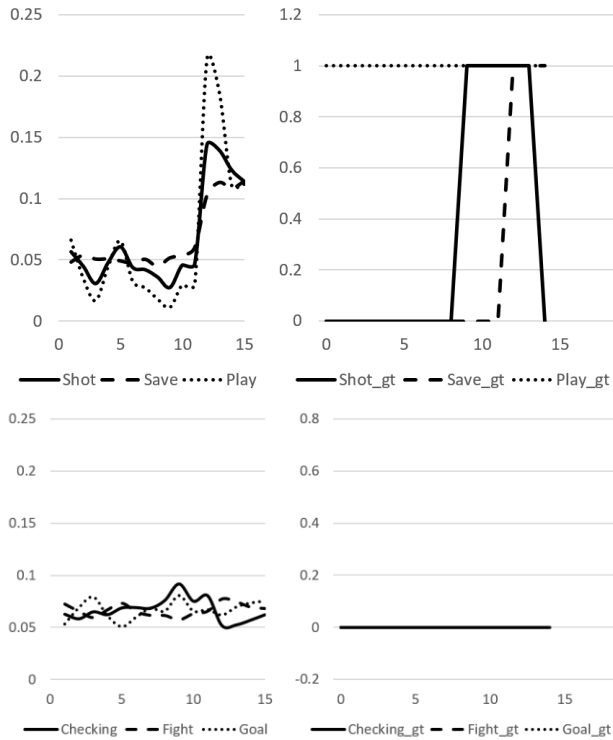

Figure 6: Visualization of temporal attention generated for separate activities. Top left, three activities happened in the 15 frames (shot and save happens at the last 5 seconds). Bottom left, activities didn't happen. Right, corresponding ground truth.

Our proposed strategy achieved state-of-the-art performance compared with previous works (Table 4). Most previous work on CelebA focused on applying spatial attention to extract important features. Our proposed strategy benefits by modeling the association between different labels. The experimental results show that our system transfers well to different multi-label classification tasks.

5.3. Limitations and Future work

Although we have shown that our system can learn separated spatial-temporal features for independent activities and achieve good performance, there are two limitations: 1. Simply flattening the feature vectors from all the clusters results in long feature vectors with redundant information. Using a better merging strategy from different clusters will be our future work. 2. Many recent works have demonstrated that non-local neural networks and i3D have good performance on activity recognition. Pre-training the feature extractor on larger single-activity recognition datasets has helped boost performance [32][7]. Our proposed feature-to-activity attention works with different feature extractors and spatial-temporal representations. Implementing a different feature extractor with our proposed system will be our next step.

| System | Accuracy | Precision (Top 10) | Recall (Top 10) |
|---|---|---|---|
| VGG Net[27] | 85 | 75 | 77 |
| MOON [22] | 91 | / | / |
| MCNN [13] | 91 | / | / |
| Walk and Learn [33] | 88 | / | / |
| FAFL [21] | 91 | / | 71 |
| DMTL [12] | 92 | / | / |
| Ours | 92 | 93 | 93 |

Table 4: Experimental results and comparison on CelebA dataset.

6. Conclusion

We designed a novel deep learning architecture for recognizing concurrent activities that outperformed state-of-the-art mechanisms in three published datasets (charades, volleyball, and hockey). We hope this paper delivers the following contribution to the society:

- A modified VLAD feature extractor and novel activity mapping layer that extracts independent features for each activity while preserving the temporal information.
- A novel triaxial self-attention that learns independent temporal associations for different activities and makes concurrent activity predictions aware of possible activity combinations.
- A novel use of the self-attention decoder that helps extract hidden associations between activities and avoid redundant weights.
- An implemented network architecture that can serve as a reference for any multi-label classification problem given sequential input.